\documentclass[lettersize,journal]{IEEEtran}
\usepackage{amsmath,amsfonts}
\usepackage{algorithmic}
\usepackage{algorithm}
\usepackage{array}
\usepackage[caption=false,font=normalsize,labelfont=sf,textfont=sf]{subfig}
\usepackage{textcomp}
\usepackage{stfloats}
\usepackage{url}
\usepackage{verbatim}
\usepackage{graphicx}
\usepackage{cite}
\usepackage{multirow}
\usepackage{array}
\usepackage{hyperref}

\def\BibTeX{{\rm B\kern-.05em{\sc i\kern-.025em b}\kern-.08em
    T\kern-.1667em\lower.7ex\hbox{E}\kern-.125emX}}
\begin{document}

%%% Commenting Commands
\newcommand{\Maggie}[1]{{\normalsize{\textbf{({\color{blue}Maggie:\ }#1)}}}}
\newcommand{\Zac}[1]{{\normalsize{\textbf{({\color{orange}Zac:\ }#1)}}}}
\newcommand{\Xiangyun}[1]{{\normalsize{\textbf{({\color{purple}Xiangyun:\ }#1)}}}}
\newcommand{\Xiang}[1]{{\normalsize{\textbf{({\color{red}Xiang:\ }#1)}}}}
\newcommand{\Talia}[1]{{\normalsize{\textbf{({\color{yellow}Talia:\ }#1)}}}}
\newcommand{\talia}[1]{{\normalsize{\textbf{({\color{yellow}Talia:\ }#1)}}}}
\newcommand{\Anvay}[1]{{\normalsize{\textbf{({\color{teal}Anvay:\ }#1)}}}}
\newcommand{\Chae}[1]{{\normalsize{\textbf{({\color{WildStrawberry}Chae:\ }#1)}}}}
\newcommand{\Saima}[1]{{\normalsize{\textbf{({\color{magenta}Saima:\ }#1)}}}}
\newcommand{\Xiaonan}[1]{{\normalsize{\textbf{({\color{green}Xiaonan:\ }#1)}}}}

\title{TALE-teller: Tendon-Actuated Linked Element Robotic Testbed for Investigating Tail Functions}

\author{Margaret J. Zhang$^{1}$,
Anvay A. Pradhan$^{1}$, 
Zachary Brei$^{1}$, 
Xiangyun Bu$^{1}$, 
Xiang Ye$^{1}$, 
Saima Jamal$^{1}$, 
Chae Woo Lim$^{1}$, 
Xiaonan Huang$^{2}$, 
Talia Y. Moore$^{1,2,3}$ %
\thanks{$^{1}$Mechanical Engineering, University of Michigan, Ann Arbor, MI {\tt\small maggiejz, anvay, breizach, xybu, yexiang, saimaj, chaewlim@umich.edu}}%
%%\thanks{$^{2}$Mechanical Engineering, Boston University, Boston, MA  {\tt\small chaewlim@bu.edu}}%
\thanks{$^{2}$Robotics, University of Michigan, Ann Arbor, MI  {\tt\small xiaonanh@umich.edu}}%
\thanks{$^{3}$Ecology and Evolutionary Biology, Museum of Zoology, University of Michigan, Ann Arbor, MI  {\tt\small taliaym@umich.edu}}%
}

\maketitle

\begin{abstract}

%200 words max
Tails serve various functions in both robotics and biology, including expression, grasping, and defense. 
The vertebrate tails associated with these functions exhibit diverse patterns of vertebral lengths, but the precise mechanisms linking form to function have not yet been established.
Vertebrate tails are complex musculoskeletal structures, making both direct experimentation and computational modeling challenging.
This paper presents Tendon-Actuated Linked-Element (TALE), a modular robotic test bed to explore how tail morphology influences function. 
By varying 3D printed bones, silicone joints, and tendon configurations, TALE can match the morphology of extant, extinct, and even theoretical tails. 
We first characterized the stiffness of our joint design empirically and in simulation before testing the hypothesis that tails with different vertebral proportions curve differently.
We then compared the maximum bending state of two common vertebrate proportions and one theoretical morphology.
Uniform bending of joints with different vertebral proportions led to substantial differences in the location of the tail tip, suggesting a significant influence on overall tail function.
Future studies can introduce more complex morphologies to establish the mechanisms of diverse tail functions.
With this foundational knowledge, we will isolate the key features underlying tail function to inform the design for robotic tails. Images and videos can be found on \href{https://www.embirlab.com/tale}{TALE's project page}.
\end{abstract}


\section{Introduction}
\label{sec:introduction}

\begin{figure}
    \centering
    \includegraphics[width=1\linewidth]{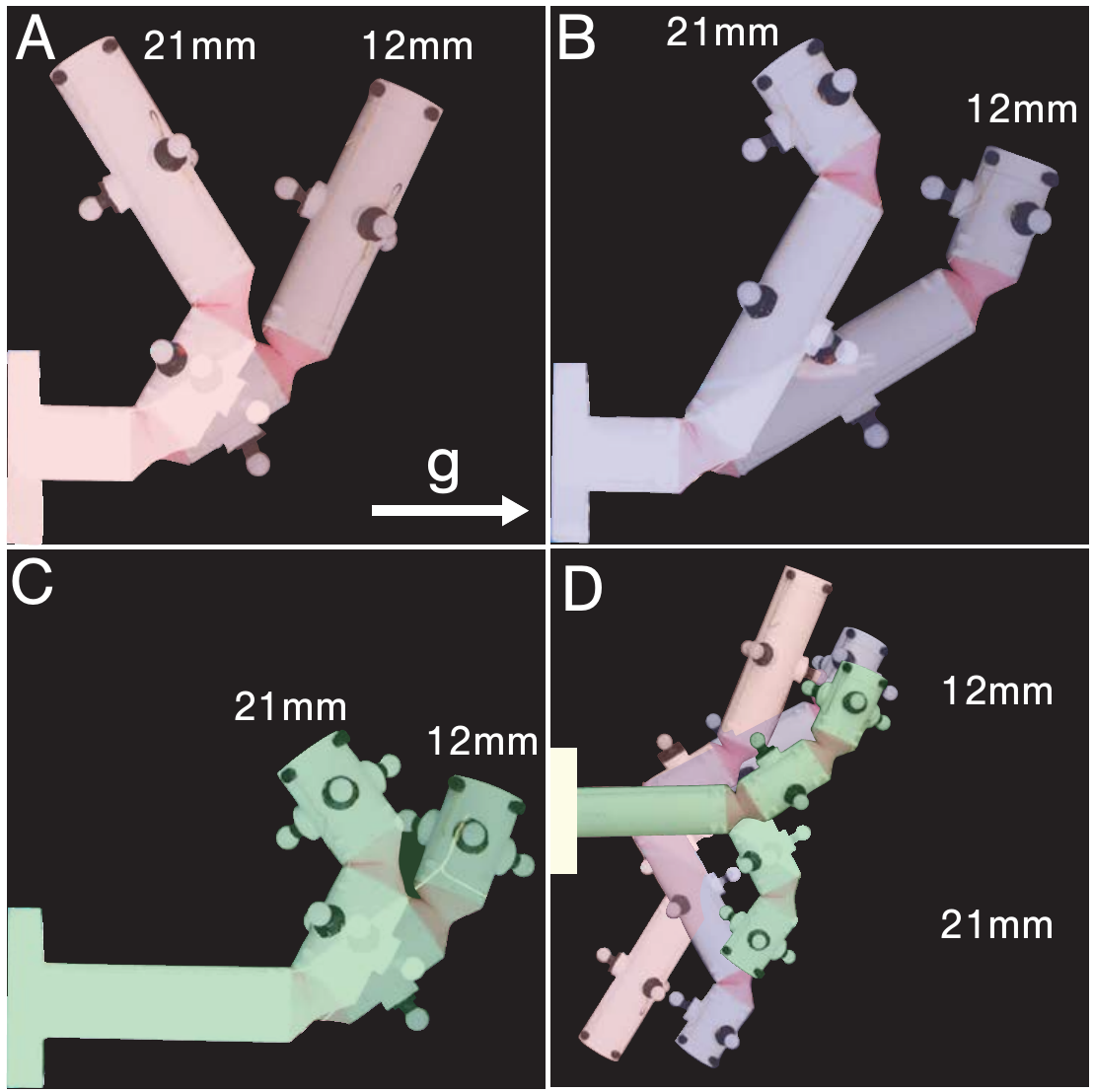}
    \caption{Tails with the same overall length but variable link lengths experience different bending behavior under the same displacement input. 
    A, B, and C show three distinct tails, each pulled to 12~mm and 21~mm of tendon displacement. 
    D shows an overlay of both displacement inputs for all three tails. 
    % Upwards deflection is 12~mm displacement, downwards deflection is 21~mm displacement. 
    Note that direction of gravity is horizontal to the right.}
    \label{fig:intro_fig}
\end{figure}

Vertebrate tails are diverse biological appendages with a wide range of functions, including communication, prehensile manipulation, and fly swatting \cite{shield2021tails,hickman1979mammalian}.
Aspects of these tail functions have informed the design of several tailed robots \cite{zhang2022versatile, libby2012tail,jusufi2010righting,liu2021dynamic}.
The complexity and functionality of robotic tails is minimal compared to their biological counterparts, partly because we do not understand the fundamental mechanics underlying most tail functions \cite{walker2006extension, shield2021tails, saab2018robotic}.
It is difficult to experimentally determine how variations in tail morphology confer specific functional advantages because biological tails are highly complex, poorly described structures \cite{schwaner2021future}.
However, recent work suggests specific tail functions are associated with distinct vertebral proportions \cite{Fu,russo2015postsacral}. 

\begin{figure*}[ht]
    \centering
    \includegraphics[width=\textwidth]{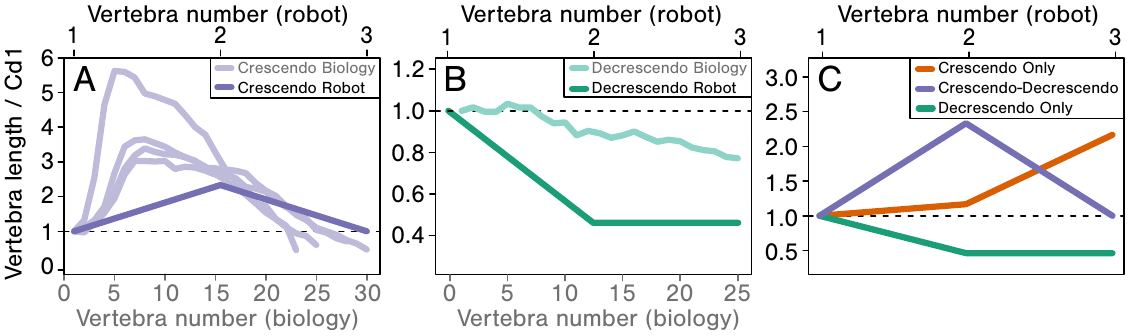}
    \caption{Our robotic platform can recreate diverse biological tail morphologies and explore the performance of theoretical morphologies.
    A) The crescendo-decrescendo robot tail is similar to the vertebral proportions of animals that use their tails as inertial appendages \cite{Fu}.
    B) The decrescendo only robot tail is similar to the tails of lizards \cite{nunez2018mesosaurus}, which are frequently used in defense for swatting \cite{carpenter1961patterns}. %\Zac{maybe specify a function here too?}
    C) All three robot tails have the same total length but differ in proportion.
    The crescendo only tail (red) explores the function of a theoretical morphology not found in vertebrate tails.
    }
    \label{fig:realtails}
\end{figure*}

Mammals using their tails primarily for inertial maneuvering---\emph{Pygathrix nemaeus}, 
\emph{Otolemur crassicaudatus},
\emph{Saimiri oerstedi}, and
\emph{Bassaricyon alleni}---exhibit an exaggerated crescendo-decrescendo pattern that may enable the tail to generate higher angular momentum to rotate their bodies \cite{Fu} (Fig. \ref{fig:realtails} A).
Lizards---\emph{Varanus rudicollis}--generally exhibit a decrescendo-only pattern and frequently use their tails for defensive strikes \cite{nunez2018mesosaurus,carpenter1961patterns} (Fig. \ref{fig:realtails} B).
A crescendo-only pattern has not been observed in any vertebrate species.
Animals with a tail tip elaboration, such as a club in ankylosaurid dinosaurs, likely formed these structures through secondary ossifications and fusions of small tail bones, rather than evolution to increase vertebral bone size consistently along the length of the tail \cite{hayashi2010function,wang2017morphological}.
Because of its absence in biology, the theoretical biomechanical performance of a crescendo-only tail has not yet been explored.

While it would be difficult to vary vertebral proportions and observe the functional consequences in biological structures, robotic physical models facilitate this kind of experimental hypothesis testing across a wide range of biological traits \cite{zhang2023launching,roberts2014testing,lauder2022robotics}.
Robotic physical models enable researchers to vary specific parameters that might be genetically correlated with other features in biology, thereby offering a highly controlled approach to identify the function of specific morphological features  \cite{ramezani2017biomimetic,ijspeert2007swimming,kim2008smooth,patek2011bouncy}.
Moreover, robotic systems allow for reproducible experimentation, testing of theoretical morphologies not found in biology, and investigation of phenomena with complex physics that are challenging for computational simulations \cite{roberts2014testing,libby2012tail,lauder2022robotics}.
This robotic approach to biomechanical research enables the discovery of biological mechanisms that can then be used to imbue robots with novel capabilities.

\subsubsection{Robot Tail Design}
Most existing robotic tails were not explicitly designed as platforms for hypothesis testing.
Based on the inertial maneuvering described in lizards \cite{libby2012tail}, many robots use a single rigid link for stabilization on a vertical surface 
\cite{liu2014bio, zhao2013controlling} or as an inertial appendage in a single plane \cite{libby2012tail,chang2011lizard,zhao2015msu, kohut2012effect, kohut2013precise} or two planes \cite{saab2019design,patel2015conical, liu2020design}.
While this rigid link tail has been sufficient to model lizards, many mammals tails exhibit complex curvatures that likely enhance three-dimensional inertial maneuvering \cite{johnson2012tail}.

Continuum robots, composed of a series of rigid links, are frequently used to confer curvature to a robotic tail.
Because the links are usually identical, continuum robots are also modular in length and link number.
Despite this modularity, few multi-link robots combine links of different sizes to confer distinct functions to different regions of the tail (but see \cite{doerfler2023hybrid}).

The structure of the joints between these links determines the capabilities of the tail.
Pin joints constrain tail motion to one plane of action \cite{doerfler2023hybrid}, but 3D motion has been achieved by combining a planar tail with a rotary base \cite{liu2020design}.
In contrast, biological tail joints are functionally omnidirectional (allowing bending in any direction and rotational twist), as most tail vertebrae lack the bony processes that limit bending and twisting in the torso \cite{jones2021autobend,organ2010structure}.

Most continuum robots are actuated by cables pulled by motors mounted at the base \cite{saab2019design,nakano2023robostrich,luo2023novel,liu2021real}.
Compared to direct-drive actuators in each joint, cable-driven systems reduce the total tail mass, can increase in link number without increasing the complexity of control, and can reach smaller sizes.
This is quite similar to the tails of mammals, which are actuated by muscles at the base of the tail that connect to elongate tendons that span several joints before inserting on a vertebra \cite{hori2011participation}.
However, the muscle-tendon architecture of mammal tails is quite complex and variable between species, with multiple overlapping sets of tendons arranged in four primary tracts along the tail \cite{Miyamae}.
%\Anvay{check citation}\cite{nakano2023robostrich,zhang2022novel}.\Maggie{These are more like tensegrity}

Considering these differences between existing tail robots and biological tails, we sought to design a robotic testbed that could accommodate different levels of biofidelity.
In particular, we focused on enabling omnidirectional, viscoelastic bending, vertebrae that can vary in size, and cable attachment sites to enable a wide range of tendon arrangements.
Furthermore, we wanted to design a tail that could be compatible with different types of actuators (i.e., rotary or linear) for generating different tail behaviors.

This paper has four main contributions.
Section \ref{sec:methods} includes the design of a modular robotic test bed (Section \ref{sec:design}).
We performed thorough characterization of the modular joint (Section \ref{sec:joint}) and found that it  mimics biological equivalents (Section \ref{sec:res_joint}).
Section \ref{sec:experiment} describes empirical testing procedures, which determined that tail deflections have high repeatability for investigating robot tail designs and animal tail morphologies (Section \ref{sec:res_exp}). 
In Section \ref{sec:discussion}, we discuss possible interpretations of our results and potential applications of this platform to further both robotics and biology.

\section{Methods}
\label{sec:methods}

We identified several key aspects of biological tails that must be sufficiently mimicked to enable hypothesis testing with a robotic physical model.
First, the number of bones must be easily altered, as there is high variation in the number of vertebrae that make up animal tails \cite{russo2015postsacral}.
The bones must also be able to vary in length; vertebrate tails include vertebrae of different lengths \cite{Fu,nunez2018mesosaurus} (Fig. \ref{fig:realtails}).
Second, the joints between bones must have 3-DoF, instead of being limited to planar motion, and have passive elastic behavior \cite{kambic2017experimental}. %\Zac{I would remove "in any DOF".}
Third, we specifically decided to mimic mammal tails, which are actuated by muscles at the base of the tail that are connected to tendons that span several joints, making the actuation of each joint non-independent.
We therefore decided to mimic tendons with cables, which can be pulled by any type of actuator.
Finally, the tendons must be routed along the tail in four tracts, as observed in mammal tails \cite{Miyamae}. 

\begin{figure}
    \centering
    \vspace{3pt}
    \includegraphics[width=1.0\linewidth]{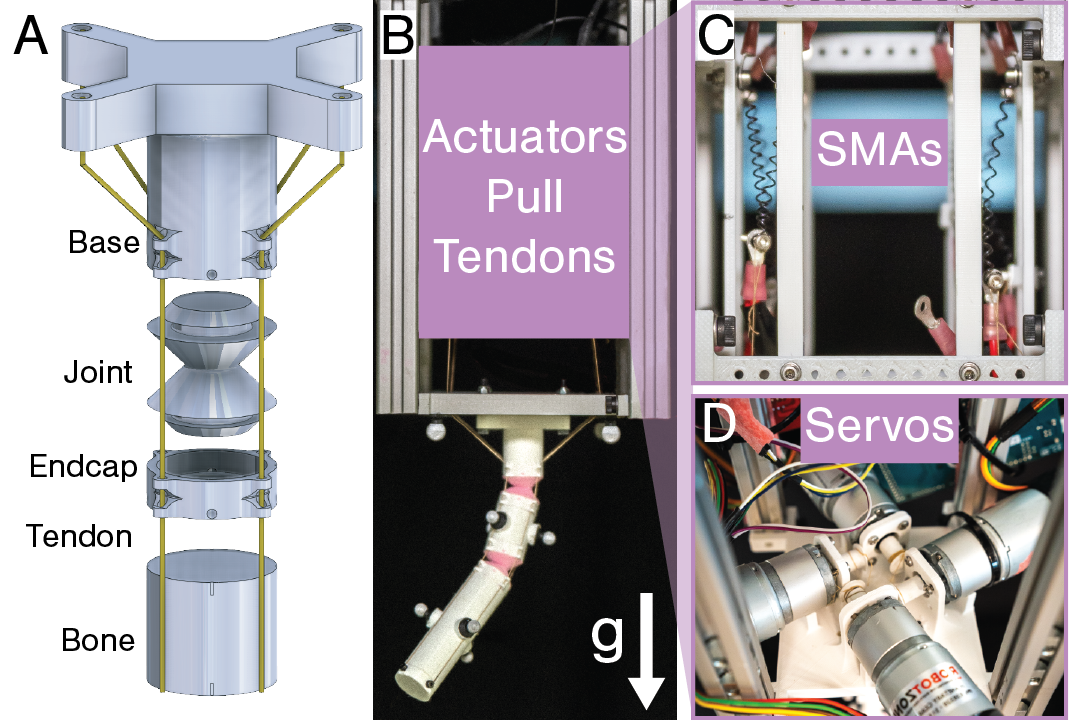}
    \caption{A) Component order of bones and joints for TALE.
    B) Completed TALE robot without actuation system.
    C) Examples of compatible actuators.}
   
    \label{fig:MotorSMA}
\end{figure}

\subsection{Tail Design, Materials, and Manufacturing}
\label{sec:design}

We designed the tail to have three main components: bones, joints, and tendons. 
We 3D printed bones using standard PLA with 20\% infill on the Prusa Mk3 (Prusa, Czech Republic). 
Each bone was composed of a center piece and two end caps, which were used to connect the bones, joints, and tendons (Fig. \ref{fig:MotorSMA} A).
This modular composition facilitated length and density changes by adjusting the infill percentage and length of the center piece. 
Each end cap also included eight loops through which the Kevlar tendons were strung.
The loops were arranged in pairs on the four dorsolateral and ventrolateral regions of each bone, which is equivalent to the tendon path of action observed in several mammal tails \cite{Miyamae}.
End caps had a uniform length of 6~mm and were attached to each end of the bone center piece using Loctite 416 instant adhesives (Henkel, Düsseldorf, Germany).

Instead of rigid joints, we opted to use soft joints, so that the material selection and geometry of the joint would determine its stiffness, and bending would not be restricted to specific planes.
We designed an hourglass-shaped mold for the joints (Fig. \ref{fig:tail_comparisons}), with a length $h$ = 12~mm, end radius $r_1$ = 10.2~mm, and neck radius $r_2$ = 4.615~mm.
The parameterized hourglass design can be modified to adjust the relative stiffness in axial rotation (twisting) compared to bending stiffness.
We used rubber silicone (Mold Max 30 Silicone Mold Rubber, Smooth-On, Macungie, PA, USA) to replicate the stiffness found in biological joints. %\Zac{take out Macungie?}
We mixed two-part silicone, vacuumed to remove air bubbles, and then cast inside the 3D-printed molds.
Once cast, we glued the joints with Loctite 416 instant adhesives onto the 3D-printed end caps and connected the bone segments to form a tail.

% The angle, length, diameter, and curvature of the joint can be manually redesigned in 3D CAD software and to be cast into each bone. \Xiang{Interchangeably using bone/link/segment. May need to decide which to use.}
% 
% All joints used in experiments for this paper are identical, thus providing constant stiffness at each joint. \Zac{I wouldn't say constant stiffness} \Xiang{do we need to mention details such as "silicone compound is vacuumed before casting to avoid air bubbles, which would affect joint properties", "joint edges are attached to the bone edges with glue to avoid unwanted joint deformation", etc}
Once assembled, we added ``tendons'' in the form of kevlar threads.
To precisely control the lengths of the tendons, we used set screws to attach the tendon to the endcap.
We threaded the tendon through the relevant loops on each bone, ultimately attaching to an actuator.
We designed the testing platform to be compatible with a wide range of actuators, including servomotors, shape memory alloys, and pneumatic artificial muscles (Fig. \ref{fig:MotorSMA} B, C, D). We used servomotors for the trials in this experiment.

\subsection{Joint Modeling and Characterization}
\label{sec:joint}

We characterized isolated joint properties to determine their mechanical properties, the variability in the manufacturing process, and to ensure that the bending behavior is similar to that of biological joints.
First, we performed Instron tests (Sec. \ref{sec:instron_experiments}) to empirically characterize the joint bending stiffness.
Second, we used the elastic modulus and Poisson's ratio for cartilage to simulate the deformation of an  equivalent biological joint in a finite element analysis (FEA) model (Sec. \ref{sec:joint_fea_model}).
Last, we developed a computational model to predict the deformation behavior of the joint (Sec. \ref{sec:joint_computational_model}). 

% an analytical model of the joint is developed in Sec. \ref{sec:joint_analytical_model}.
% Then, the joint was simulated using Finite Element Analysis (FEA) experiments that are detailed in Sec. \ref{sec:joint_fea_model}.
% Finally, both the joints and tendons were experimentally characterized using an Instron testing machine (68SC-5, Instron, USA), which is detailed in Sec. \ref{sec:instron_experiments}.
% \Zac{order needs to be updated}

\subsubsection{Experimental Characterization}
\label{sec:instron_experiments}
% Results section will have summary of joint modeling and testing
We performed tensile tests to obtain the tail joint force-displacement curves, compute the elastic modulus, and characterize the variability in the manufacturing process (Supplementary Video 2).
For this test, we used three identical tail segment samples that consisted of a bone-endcap-joint-endcap configuration.
We fixed the testing sample in a custom 3D-printed fixture, which was then mounted on the Instron testing machine (68SC-5, Instron, USA). 
This setup included a 100~N load cell with a resolution of 0.0001~kN and a customized Stainless Steel 304 gripper.
The 3D-printed fixture was a hollow cuboid frame (40~mm x 24~mm x 75~mm) with clearance to observe the deflection of the sample.
We tied one end of a Kevlar string to a slip pin in the top grip of the Instron machine, with the other end tied to the bottom endcap. 
This setup ensured that the tendon pulled the sample with the same motion as in the complete TALE robot.
We performed six trials for each of three samples by displacing the string at 10~mm/min. 
We measured the tendon pulling forces for displacements ranging from 0~mm to 9~mm, in 1~mm increments.

% The force-displacement data was recorded using Bluehill Universal software and the mean force-displacement curve was shown for each joint in Fig. \ref{}. 
% Data analysis included calculating the mean force-displacement curve and evaluating the variability of joint stiffness across trials.\Xiang{not sure the two comments above are necessary.} \Zac{can put this in results when the figure is referenced}

\subsubsection{Finite Element Analysis}
\label{sec:joint_fea_model}
We selected the joint material to match a biological vertebratal joint. 
To show that the selected joint design fell within a biological range, we used FEA to simulate the bending of the same tail segment used in the Instron testing (Sec. \ref{sec:instron_experiments}).
Biological intervertebral discs are made of cartilage, so we applied the material properties of cartilage (elastic modulus and Poisson's ratio) to the joint. 
% The Instron tests are are detailed in Section \ref{sec:instron_experiments}.
% The tail segment was composed of the same elements used in the Instron tensile tests, a joint connected to a rigid bone segment and a rigid end cap.
We imported the single joint segment into Altair Hypermesh 2022.1 (Altair, Troy, MI, USA). %\Xiang{what system?}
% To accurately simulate tail bending in a manner close to that of the natural world, the material properties of animal tail cartilage were used for the soft joint in the FEA model. \Zac{I am confused by this sentence. Don't we know what the actual properties of the joint material are? Weren't those used in the FEA simulation?}
% \Xiangyun{Mold Max 30 is a hyper-elastic material, there is no young's modulus provided for it. We agreed on using a similar material in the natural world to run FEA under linear mode}

We used the printed tail bone material properties: standard PLA filament, with a density of $1.24$~g/cm$^3$, a Poisson's ratio of 0.36, and a elastic modulus of $3.6$~GPa \cite{pepelnjak2020altering}.
We found a range of cartilage values from the literature, so we modeled the extreme values of the range to bound the model predictions \cite{mansour2003biomechanics}.
For all joints, we used a density of $1.07$~g/cm$^3$, but varied the Poisson's ratio between $0.30-0.40$, and an elastic modulus of $0.45-0.80$~MPa to define the lower and upper bounds, respectively.
The displacement constraints experienced by the joint during Instron testing informed equivalent boundary conditions in the FEA model. 
We rigidly fixed the bone, which accordingly fixed the endcap and the surface of the flexible joint in contact with the bone.
We applied a 9~mm displacement constraint to the location of tendon attachment in the direction of the tendon pull in Instron testing.
Because the displacements in the joint pulling tests were within the elastic range of the materials, we ran the FEA under a linear static mode.

\subsubsection{Computational Modeling}
\label{sec:joint_computational_model}

\begin{figure*}
    \centering
    \includegraphics[width=1.0\textwidth]{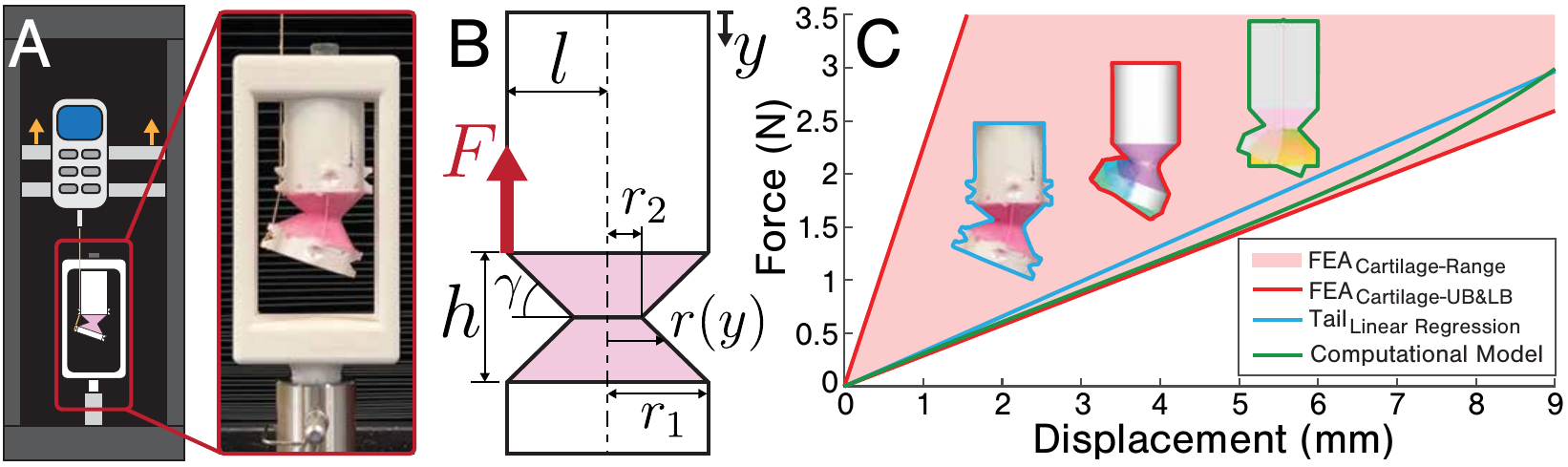}
    \caption{A) Illustration of Instron testing setup.
    B) Schematic of one joint. 
    Note that $l$ is the moment arm and $\gamma$ is the angle of the joint with respect to horizontal. 
    Both values change as the joint deforms. 
    $h=12$~mm, 
    $r_1=10.2$~mm, 
    $r_2=4.615$~mm, 
    $l=r_2+\sqrt{\left( r_1-r_2 \right)^2+\left( h/2 \right)^2}\cos\gamma$. 
    C) A combined plot showing the bending behavior and force-displacement curves of the joint during Instron testing (cyan), FEA simulation (red), and computational model deformation (green).}
   
    \label{fig:tail_comparisons}
\end{figure*}

We constructed a computational reduced order model to simulate the deformation of the tail joint segment used in the Instron testing.
Our computational model can be extended to simulate and predict the behavior of a multi-jointed tail.
This simulation uses a finite element-like formulation in which local stiffness matrices are computed by taking the second variation of a potential function before being summed into a global stiffness matrix.
Relevant parameter definitions are presented in Fig. \ref{fig:tail_comparisons} B.

We represented the joint as a three node system connected by two bars in 3D.
Nodes possessed only translational degrees of freedom.
For a node $\underline{x}_i$, its position was given by $\underline{x}_i=\{x_{i1},x_{i2},x_{i3}\}$.
After deformation, the same node location was equal to the original position plus a displacement $\underline{u}_i$ where $\underline{u}_i=\{u_{i1},u_{i2},u_{i3}\}$.
Therefore, the future node location $\underline{x}_i'$ can be written as $\underline{x}_i'=\underline{x}_i+\underline{u}_i$.
Note that $\underline{u}_i$ was initially unknown.

In the simulation, bars connect any two nodes $\underline{x}_i$ and $\underline{x}_j$.
For the tail joint, bars were aligned with the centerlines of each half of the joint.
Bar elements follow a discrete elastic rod formulation in which the element only captures stretching or compression.
Therefore, these bars provide a measure of the axial stiffness of the joint.
The potential function for bars is given in Eq. \ref{eq:barPotential}.

\begin{equation}
    \Pi_{bar} = \frac{1}{2} \frac{EA}{L_0} \left( L-L_0 \right)^2
    \label{eq:barPotential}
\end{equation}
where
\begin{equation}
    L_0 = \Vert \underline{x}_i-\underline{x}_j \Vert
    \label{eq:L0}
\end{equation}
\begin{equation}
    L = \Vert \underline{x}'_i-\underline{x}'_j \Vert
    \label{eq:L}
\end{equation}

Note that $E$ is the elastic modulus, $A$ is cross-sectional area, and $\Vert * \Vert$ refers to the 2-norm.
$E$ is determined empirically while $A=A(y)=\pi\left(r(y)\right)^2$ varies as a function of the radius of the joint $r(y)$.
The radius $r(y)$ changes linearly along $y$.
Since the joint has two bilaterally symmetric sections, it was divided into two halves with radii given by $r_I(y)$ and $r_{II}(y)$ as shown in Eq. \ref{eq:rI} \& \ref{eq:rII}.
\begin{equation}
    \label{eq:rI}
    r_I(y) = \left( \frac{r_2-r_1}{h/2} \right) \left( y-\frac{h}{2} \right) + r_2 \;\;\;\;\; \forall y\in \left[ 0,\;\frac{h}{2} \right]
\end{equation}
\begin{equation}
    \label{eq:rII}
    r_{II}(y) = \left( \frac{r_1-r_2}{h/2} \right) \left( y-\frac{h}{2} \right) + r_2 \;\;\;\;\; \forall y\in \left[\frac{h}{2},\;h \right]
\end{equation}
To capture bending stiffness of the joint, a rotational spring was included between all three nodes $\underline{x}_i$, $\underline{x}_j$, and $\underline{x}_k$.
The potential function for the rotational spring is given in Eq. \ref{eq:sprPotential}.

\begin{equation}
    \Pi_{spr} = \frac{1}{2} k_\theta \left( \theta-\theta_0 \right)^2
    \label{eq:sprPotential}
\end{equation}
where
\begin{equation}
    \theta_0 = \arccos \left( \frac{\left(\underline{x}_i-\underline{x}_j \right) \cdot \left( \underline{x}_k-\underline{x}_j \right)}{\Vert \underline{x}_i-\underline{x}_j \Vert  \Vert \underline{x}_k-\underline{x}_j \Vert} \right)
    \label{eq:theta0}
\end{equation}
\begin{equation}
    \theta = \arccos \left( \frac{\left(\underline{x}'_i-\underline{x}'_j \right) \cdot \left( \underline{x}'_k-\underline{x}'_j \right)}{\Vert \underline{x}'_i-\underline{x}'_j \Vert  \Vert \underline{x}'_k-\underline{x}'_j \Vert} \right)
    \label{eq:theta}
\end{equation}

Note that $\theta - \theta_0$ represents the change in angle of the joint and that $k_\theta$ is the equivalent spring stiffness of the joint, which was determined empirically using the Instron test.
For axial stiffness, we derived the elastic modulus from uniaxial material testing of the silicone joint.
For bending (spring) stiffness, we fitted a linear curve to the cantilever test of the silicone joint.

Once potentials were defined, the second derivative was computed with respect to the displacements (Eq. \ref{eq:secondVariation}).
\begin{equation}
    \delta = \frac{\partial^2\Pi}{\partial u_{ij} \partial u_{mn} }
    \label{eq:secondVariation}
\end{equation}
For the bar elements, $\delta = \left[ 6\times6 \right]$ and for the rotational springs, $\delta = \left[ 9\times9 \right]$.
Computation of these derivatives was done using multivariate finite differences (Eq. \ref{eq:finDiffDiag} \& \ref{eq:finDiffOffDiag}). 
For a term $x_{ij}$, let $\underline{x}_{\widehat{ij}}$ represent all terms in $\underline{x}$ except $x_{ij}$. 
Let $\Delta \ll 1$.

\begin{flalign}
    \label{eq:finDiffDiag}
     \forall \;\; ij=mn   \\ 
     \frac{\partial^2\Pi}{\partial u_{ij}^2} &= \frac{\Pi\left( \notag x_{ij}+\Delta,\underline{x}_{\widehat{ij}},\underline{u}=0 \right)}{\Delta^2} & \\ \notag
     & -\frac{2\Pi\left( x_{ij},\underline{x}_{\widehat{ij}},\underline{u}=0 \right)}{\Delta^2} & \\ \notag
     & +\frac{\Pi\left( x_{ij}-\Delta,\underline{x}_{\widehat{ij}},\underline{u}=0 \right)}{\Delta^2} \notag &
\end{flalign}

\begin{flalign}
    \label{eq:finDiffOffDiag}
    & \forall \;\; ij\neq mn & \\ 
    & \frac{\partial^2\Pi}{\partial u_{ij}\partial u_{mn}} = \notag
    \frac{\Pi\left( x_{ij}+\Delta,x_{mn}+\Delta,\underline{x}_{\widehat{ij} \& \widehat{mn}},\underline{u}=0 \right)}{4\Delta^2} & \\ \notag
    &\;\;\;\;\;\;\;\;\;\;\;\;\;\;\;\;  -\frac{\Pi\left( x_{ij}+\Delta,x_{mn}-\Delta,\underline{x}_{\widehat{ij} \& \widehat{mn}},\underline{u}=0 \right)}{4\Delta^2} & \\ \notag
    &\;\;\;\;\;\;\;\;\;\;\;\;\;\;\;\;  -\frac{\Pi\left( x_{ij}-\Delta,x_{mn}+\Delta,\underline{x}_{\widehat{ij} \& \widehat{mn}},\underline{u}=0 \right)}{4\Delta^2} & \\ \notag
    &\;\;\;\;\;\;\;\;\;\;\;\;\;\;\;\;  +\frac{\Pi\left( x_{ij}-\Delta,x_{mn}-\Delta,\underline{x}_{\widehat{ij} \& \widehat{mn}},\underline{u}=0 \right)}{4\Delta^2} \notag &
\end{flalign}

For loading, pulling of the tendon induced a compressive point load on the outer edge of the joint.
This load was equivalently represented as a moment, $M=Fl$, applied to the free end of the joint.
The moment was then decomposed into two forces acting on the furthest node, one in the direction of the pulling $F_{\parallel}=F$ and the other orthogonal to the pulling $F_{\perp}=\frac{2M}{h}$.
Due to the large geometric non-linearity during joint bending, we implemented a non-linear Euler method to slowly increment the load and better approximate the final position of the tail.
While not as accurate as other non-linear solvers, using the Euler method enabled fast computation, making the computational model versatile for testing many loading scenarios.

% \subsubsection{Experimental Characterization of Tendons}
% \Zac{combine with joint experimental section}
% \Zac{figure for this or just report the numbers? could put figure in appendix or on website. Also, are there not numbers from the manufacturer and that is why we tested this?}\Xiang{Chae and Anvay worked on this. Not sure if they took setup photos and even if they do it will be an individual figure, which may not be a good idea. This section can be added to the end of 3) above? Data from manufacturer is not clear and has conditions.}

% \subsection{Tail Simulation Modeling}

% Each variation of the physical tail model was recreated using ArborSim \cite{fu2024arborsim} to create a multi-jointed tendon-driven system in OpenSim. 
% The STL files for the bone segments and joints were imported into OpenSim and the joints had physical landmarks assigned for the length and directions of each joint. \Zac{what is length here?} 
% Each of the four tendons in the simulation were created by back calculating a hypothetical muscle that would contract the same amount that SMAs of the physical model do. 
% Each tendon is also created to be inextensible to mimic the inextensible Kevlar thread. 
% Data from the Instron based characterization of the system is used to define the joint motion and stiffness in the OpenSim model. 
% Motion and position of the 3DTail is verified using this model through simulation of full activation of the muscles.

\subsection{Testing the effect of tail morphology on bending}
\label{sec:experiment}

We demonstrate the use of this robotic physical model by investigating the effect of different tail morphologies, specifically variations in bone segment length, on the bending behavior of a tail.
To do so, we chose a fixed total length for the tail and tested three different morphologies composed of three bones (two short, 18~mm, and one long, 60~mm) in three configurations (short-short-long, short-long-short, long-short-short) (Fig. \ref{fig:realtails} (c)).
We designed the first bone to be integrated with the base and immobile for all morphologies, resulting in three elastic joints.
When combined, all tails were 150~mm in total length.

We used two pulling configurations, either one motor pulling alone or two motors pulling together, where the two motor configuration used sequential pairs of motors to pull on the corresponding tendons (Fig. \ref{fig:intro_fig}).
This resulted in eight pulling directions per tail. 
% For each direction, the two one-joint tails had a tendon pull distance of 12~mm, which was the maximum displacement allowed by the joint.
% The three two-joint tails had two tendon pull distances for each direction. 
For each pulling configuration, we tested two tendon contraction distances: 12~mm and 21~mm (maximum tail deflection).
% The first pull distance matched the one-joint tails to enable a direct comparison, and the second pull distance was 21~mm.
% The second distance corresponded to the maximum displacement of the two-joint tails. 
When actuating the tail, we only attached the tendon(s) being pulled to the last end cap on the tail. 
We performed tests with tails oriented towards the direction of gravity, so as not to bias tail motion in any direction (Fig. \ref{fig:MotorSMA}).
% Note that the only tendons connected to the tail were the ones actively being used to actuate the tail, meaning the opposing tendons were disconnected. \Zac{phrase better}

We recorded the position of each bone using a motion capture (mocap) system with ten infrared cameras (Flex 3, Optitrack, USA).
We placed three markers to each moving bone and to the base plate of the testing platform, which served as the ground plane of the system.
For each bone, we placed the three markers so that the center of mass of the bone coincided with the centroid of the markers.
We recorded marker position data using Motive 2.3.1, and the data was processed in MATLAB. 
Videos of these experiments can be found on \href{https://www.embirlab.com/tale}{TALE's project page}.
% For the mocap measurements, the residual error of the marker position was 0.6 $\pm$ 0.1~mm across all trials. 
% The statistical significance of the location of the end effector is analyzed using a two-way ANOVA test. 
\section{Results} 
 \label{sec:results} 

 \subsection {Joint Characterization} 
 \label {sec:res_joint} 
 % \Maggie{this section should report more on the empirical outputs } 
   
 % We experimentally characterized the elastic joint using an Instron testing machine, as shown in Fig . \ref{fig:tail_comparisons}B. 
We compared the empirical Instron results from each of the three joint samples by first computing the average standard deviation at each displacement for each sample.
Sample 1 had an average standard deviation of 0.0390~N, Sample 2 had 0.0411~N, Sample 3 had 0.0405~N.
Because the values were so similar, we fit a linear regression to all eighteen trials, which resulted in an R-squared value of 0.994 (Fig. \ref{fig:tail_comparisons} C cyan). 
% Due to the low resolution of the Instron testing machine, we offset the linear regression curve so that the onset of data collection would originate at zero.

The FEA models output force-displacement curves representing the upper and lower boundaries of cartilage behavior (Fig. \ref{fig:tail_comparisons} C red).
The empirical results were within this bounded area, very close to the lower bound.
At 3~mm displacement, the nodal force in the simulation was 0.8567~N, which closely matched the 0.8143~N recorded by the Instron pulling test at the same displacement.

The computational model required a force input, as opposed to the displacement input in Instron testing and FEA.
Given the linearly elastic response of the joint during Instron testing, the computational model related a given displacement with its associated force.
With this input, the simulation predictions of deformation for each individual joint were within 4\% of the Instron test (Fig. \ref{fig:tail_comparisons} C).

 \subsection{Hardware Results} 
 \label{sec:res_exp} 
We recorded eleven trials for each of the eight distinct actuations for all three morphologies. 
However, due to the markers being close together, some marker identities were not discernible by the mocap software, resulting in a few unusable trials.
The minimum number of trials for any experimental condition was 8 (for Tail 1, 21~mm contraction, 2 motor actuation) and the maximum number of trials was 11.

By comparing multiple trials of each experimental condition (same tail configuration, same displacement, same motor), we examined the variation in tail behavior due to variation within each motor. 
We first computed the mean location of the end effector in 3D space. 
For each trial, we calculated the distance between the trial and this calculated mean. 
The standard deviation of this distance for motor 1 was 0.46~mm, 
motor 2 0.17~mm, 
motor 3 0.17~mm, and 
motor 4 0.20~mm.

By comparing all trials of a tail configuration and displacement, we examined the variation in tail behavior due to differences between motors.
We first collapsed all the final bone centroid positions onto a common plane defined by the central axis of the tail and a radial axis parallel to the base.
We then computed the standard deviation of the pair-wise differences of each joint position for each testing condition (Table \ref{tab:data_table}).
Because joint 1 was connected to the stationary base, its location never changed and is not reported.
In all cases, the variation in joint positions increased along the length of the tail; the standard deviation of joint 3 was always more than twice the standard deviation of joint 2.
Furthermore, the variability of each motor compounded when both were bending the tail at the same time: two motor trials always had a higher standard deviation than one motor trials. 
% This is due to the higher torque applied when two motors are actuated, which led to a higher bending angle.

 \begin{table}[h]
    \centering
        \caption{ANOVA table showing the p-values for perpendicular (left) and radial (right) distances with 12~mm displacements.
    The lower left side of each subtable shows p-values for one motor actuations.
    The upper right side shows p-values for two motor actuations.}
    \begin{tabular}{c|ccc|ccc}
        12~mm & \textbf{T1d} & \textbf{T2d} & \textbf{T3d}  & \textbf{T1r} & \textbf{T2r} & \textbf{T3r}\\
        \hline
        \textbf{T1} &  -  &  0* &  0* &  -  &  1.1e-32* &  0* \\
        \textbf{T2} &  4.2e-43*  &  - & 0.15 &  0*  & -  & 7.1e-22* \\
        \textbf{T3} &  0*  &  7.0e-3* &-&  0*  &  4.8e-29* &-\\
    \end{tabular}
    \label{tab:anova12}
\end{table}

\begin{table}[h]
    \centering
        \caption{ANOVA table showing the p-values for perpendicular (left) and radial (right) distances with 21~mm displacements.
    The lower left side shows p-values for one motor actuations.
    The upper right side shows p-values for two motor actuations.}
    \begin{tabular}{c|ccc|ccc}
        21~mm & \textbf{T1d} & \textbf{T2d} & \textbf{T3d}& \textbf{T1r} & \textbf{T2r} & \textbf{T3r}  \\
        \hline
        \textbf{T1} &  -  &  0* &  0* &  -  &  0.2 &  0* \\
        \textbf{T2} &  0*  &  - & 2.6e-10*&  1.2e-81*  &  - & 0* \\
        \textbf{T3} &  0*  &  1.1e-9* &-&  7.0e-36*  &  0* &-\\
    \end{tabular}
    \label{tab:anova21}
\end{table}

\begin{figure} 
 \centering 
 \includegraphics[width=1\columnwidth]{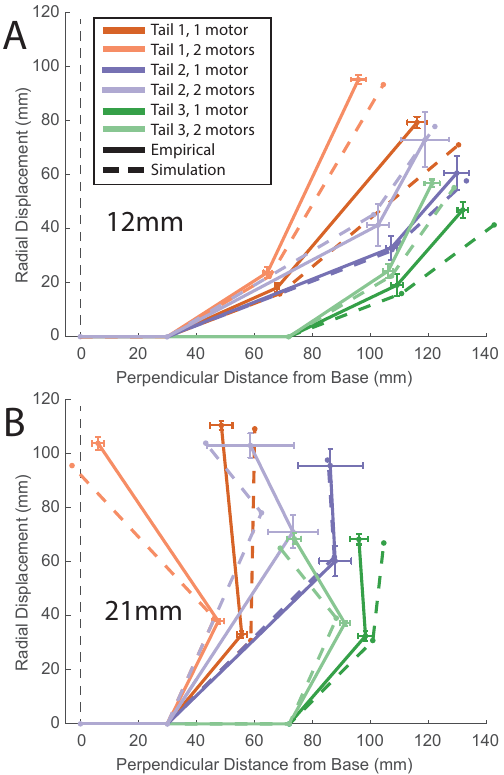} 
 \caption{
 Empirical and computational model results of tail bending for three distinct tails. 
 Solid lines indicate data from empirical tests and dashed lines indicate simulated orientations from the computational model.
 Darker lines indicate one motor actuations and lighter lines indicate two motor actuations.
 Measurements were taken in 3D using motion capture and then collapsed into a single plane defined by the central axis of the tail and a radial axis originating from the center of the tail, parallel to the base. 
 A) Tendons displaced 12~mm in testing and in simulation.
 B) Tendons displaced 21~mm in testing and in simulation.
}
 \label{fig:exp_tail_displacement}
\end{figure}

The overall bending behavior of a tail was determined by both the bone segment length and the number of tendons actuating the tail (Fig. \ref{fig:exp_tail_displacement}). 
We used one-way ANOVA tests with Tukey Honest Significant Differences to compare the perpendicular distances from the base and the radial displacement between different tail configurations.
For tails with a 12~mm contraction with one motor, there was a significant effect of group on result (F(2, 117) = 258.72, p = 4.215e-43, alpha = 0.05, Table \ref{tab:anova12} left, lower left). 
%Between Tail 1 and Tail 2, p = 5.727e-36. Between Tail 1 and Tail 3, p = 0. Between Tail 2 and Tail 3, p = 0.007.
For tails with a 12~mm contraction with two motors, there was a significant effect of group on result (F(2, 117) = 269.54, p = 6.172e-44, alpha = 0.05, Table \ref{tab:anova12} left, upper right). 
%Between Tail 1 and Tail 2, p = 0. Between Tail 1 and Tail 3, p = 0. Between Tail 2 and Tail 3, p = 0.156.
For tails with a 24~mm contraction with one motor, there was a significant effect of group on result (F(2, 117) = 523.98, p = 3.369e-58, alpha = 0.05, Table \ref{tab:anova21} left, lower left). 
%Between Tail 1 and Tail 2, p = 0. Between Tail 1 and Tail 3, p = 0. Between Tail 2 and Tail 3, p = 1.097e-9.
For tails with a 24~mm contraction with two motors, there was a significant effect of group on result (F(2, 117) = 551.42, p = 2.427e-59, alpha = 0.05, alpha = 0.05, Table \ref{tab:anova21} left, upper right). 
%Between Tail 1 and Tail 2, p = 0. Between Tail 1 and Tail 3, p = 0. Between Tail 2 and Tail 3, p = 2.555e-10.

% The following are one-way ANOVA tests with Tukey Honest Significant Differences on the radial displacements.
For tails with a 12~mm contraction with one motor, there was a significant effect of group on result (F(2, 117) = 623.35, p = 4.054e-63, alpha = 0.05, Table \ref{tab:anova12} right, lower left). 
% Between Tail 1 and Tail 2, p = 0. Between Tail 1 and Tail 3, p = 0. Between Tail 2 and Tail 3, p = 4.838e-29.
For tails with a 12~mm contraction with two motors, there was a significant effect of group on result (F(2, 117) = 411.4, p = 1.165e-53, alpha = 0.05, Table \ref{tab:anova12} right,  upper right). 
% Between Tail 1 and Tail 2, p = 1.132e-32. Between Tail 1 and Tail 3, p = 0. Between Tail 2 and Tail 3, p = 0.7.134e-22.
For tails with a 24~mm contraction with one motor, there was a significant effect of group on result (F(2, 117) = 1355.8, p = 1.180e-81, alpha = 0.05, Table \ref{tab:anova21} right, lower left). 
% Between Tail 1 and Tail 2, p = 7.023e-36. Between Tail 1 and Tail 3, p = 0. Between Tail 2 and Tail 3, p = 0.
For tails with a 24~mm contraction with two motors, there was a significant effect of group on result (F(2, 114) = 1580.26, p = 7.589e-84, alpha = 0.05, Table \ref{tab:anova21} right, upper right). 

\begin{table}[ht]
\centering
\caption{Standard deviation of the pairwise differences when comparing tail morphologies actuated by one or two motors}
\label{tab:data_table}
\begin{tabular}{|c|c|c|c|c|c|}
\hline
\multirow{2}{*}{\bf{Tail \#}} & \multirow{2}{*}{\bf{Contract}} & \multirow{2}{*}{\bf{\shortstack{\# \\ Motors}}} & \multirow{2}{*}{\bf{\shortstack{\# \\ Trials}}} & \multicolumn{2}{c|}{\bf{Motor STD (mm)}} \\
\cline{5-6}
& & & & \multicolumn{1}{c|}{\bf{Joint 2}} & \multicolumn{1}{c|}{\bf{Joint 3}} \\
\hline
\multirow{4}{*}{Tail 1} & \multirow{2}{*}{12 mm} & 1 & 40 & 5.55 & 24.08 \\
\cline{3-6}
& & 2 & 40 & 7.28 & 29.01 \\
\cline{2-6}
& \multirow{2}{*}{21 mm} & 1 & 40 & 10.09 & 33.59 \\
\cline{3-6}
& & 2 & 39 & 11.59 & 31.57 \\
\hline
\multirow{4}{*}{Tail 2} & \multirow{2}{*}{12 mm} & 1 & 41 & 5.95 & 14.42 \\
\cline{3-6}
& & 2 & 40 & 7.40 & 17.44 \\
\cline{2-6}
& \multirow{2}{*}{21 mm} & 1 & 41 & 9.98 & 20.82 \\
\cline{3-6}
& & 2 & 44 & 11.42 & 21.03 \\
\hline
\multirow{4}{*}{Tail 3} & \multirow{2}{*}{12 mm} & 1 & 43 & 10.38 & 18.93 \\
\cline{3-6}
& & 2 & 43 & 12.86 & 22.64 \\
\cline{2-6}
& \multirow{2}{*}{21 mm} & 1 & 42 & 18.97 & 28.98 \\
\cline{3-6}
& & 2 & 45 & 22.02 & 31.48 \\
\hline
\end{tabular}
\end{table}

Assuming all joints within a tail deform equally under a given tendon pull, we can use the simulation to predict the final rest positions of each bone of the tail (Fig. \ref{fig:exp_tail_displacement}A \& \ref{fig:exp_tail_displacement}B dashed).
We compared the simulation predictions to the empiricial results by calculating the root mean squared (RMS) errors of the final tail joint positions. 
The normalized RMS errors were below 6\% for all joints and all motor combinations.
For one motor pulling 12~mm, the normalized RMS errors were 4.44\%, 5.74\%, 5.8\% for tails 1, 2, and 3, respectively.
Notably, the error for two motors, each pulling at 12~mm was simlar in magnitude: 3.51\%, 5.18\%, 5.32\%.
For one motor pulling 21~mm, the normalized RMS errors were 4.12\%, 4.24\%, 4.29\%.
Similarly, the two motor pulls at 21~mm were also of the same magnitude: 4.55\%, 3.97\%, 4.05\%.

\section{Discussion}
\label{sec:discussion}
Biological form-function relationships have been useful sources of inspiration for robot design \cite{cho2016biomimetic}.
However, not all biological structures have been sufficiently characterized to inform robot design.
Without understanding the mechanisms underlying form-function relationships, designers could mimic a biological structure that has sub-optimal performance, is overly complex, or does not match the intended function \cite{fish2014evolution}.
Thus, when using bio-inspiration to guide robot design, it is crucial to consider both biologically equivalent and theoretical morphologies \cite{lauder2022robotics}. 

This paper presents the design of a modular robotic platform for understanding how vertebral morphology affects tail function in 3D space.
The key aspects of the design that make it amenable to testing biological hypotheses are:
modular bones,
soft joints with predictable and reliable deformations,
tendon-driven actuation,
and biologically informed location of tendon paths of action.
Importantly, the thorough characterization of the molded joints reveals that the manufacturing process has low variability, even when performed by different people.
Therefore, we are confident that variation in results is primarily due to differences in morphologies, rather than inconsistencies in the manufacturing process.

We also demonstrated how TALE can be used to empirically test a biomechanical hypothesis by examining the effect of bone proportions on overall bending.
% We compared the joint locations of three tails with the same total length, but with different bone lengths and bone arrangements.
% Because both joints were maximally bent at a tendon contraction of 21~mm in all tails, all joint angles were the same, so the differences in shape are the result of the changes in proportion.
The different tail morphologies resulted in significant differences in perpendicular distance from the base and in radial displacement from the central axis. 
A tail that can achieve a low perpendicular distance from the base at the tip is better at reaching towards the hypothetical torso, which likely aids in fly swatting \cite{mooring2007insect}.
The increased radial range could also potentially enhance the performance of tail functions, such as defensive striking, or ``whipping'' \cite{carpenter1961patterns}.
In each of our experiments, the theoretical crescendo-only tail has the highest values in both of these metrics.
To our knowledge, there are no vertebrate tails, extant or extinct, that exhibit a crescendo without a subsequent decrescendo.
Despite the absence of this morphology in nature, our experiments demonstrate its advantages for potential robotic tail functions.

% Understanding the range of motion of vertebral joints is essential to understanding the mechanics of locomotion, respiration, and even communication behaviors.
% Current computational approaches to determining vertebral joint ranges of motion rely on vertebral processes that are not present on all vertebrae and require significant computational power, making it difficult to model highly jointed morphologies, such as tails \cite{Fu}~(some \emph{Mesosaurus} fossils have over 60 tail vertebrae \cite{nunez2018mesosaurus}). 

%\Maggie{THIS IS THE ADDED 3 SENTENCES ABOUT ROBOTS}

Future designs will be able to leverage the biological tail mechanisms revealed by this robot to selectively integrate tail morphology into tail-like robots to directly improve the system's intended function. 
The design and use of the TALE robot not only provides deeper insight into how a biological tail might function, but also provides inspiration for possible advancements in robotics.

% The amount of change in orientation depends on the number of joints and not the distribution of link length. 
% The larger the number of joints, the larger the change in orientation of the tail end-effector.

% We designed this modular robotic testing platform so researchers can systematically explore different robotic tail designs and use that information to guide the design of robotic tails tailored for specific functions. 

\subsection{Future Work}
We simplified the shapes of the bones to validate the fundamental operation of the TALE robot, but future studies can replace the simplified bones with more realistic vertebral shapes.
Most caudal vertebrae lack bony processes (projections that extend from the bone centrum), so the range of motion is likely limited primarily by the material properties of soft tissue structures (e.g., intervertebral discs and ligaments), as in our TALE robot.
On the other hand, the range of motion of thoracic and lumbar vertebrae is largely dictated by contact between neighboring processes, or projections from the centrum \cite{jones2021autobend}.
In addition to the existing digital models of these interactions, it would be interesting to compare the predictions to TALE robots composed of the same 3D vertebral morphologies.

Furthermore, the TALE robot tendon configuration can be easily adjusted to mimic the muscle-tendon architecture of biological tails \cite{Miyamae}.
This can help us understand how different muscle actuation patterns result in distinct behaviors, can aid in uncovering patterns in musculoskeletal evolution, and can help us determine the simplest robot design that can achieve a desired functionality.

We demonstrated how the TALE robot can be used to understand the effect of vertebral proportion on maximum curvature, which is a static metric.
The platform can be modified to model more dynamic tail behaviors, such as fly swatting or inertial maneuvering, by developing more sophisticated motor controllers and integrating sensors.
% However, because inertial maneuvering operates on the concept of total conserved angular momentum for a body not in contact with the ground, the rotational motors in the current design would likely bias the results.
% This specific function requires actuation sources that do not contribute angular momentum to the robot, making linear actuators (e.g., shape memory alloys (SMA) or pneumatic artificial muscles) more suitable. 

%Copypasta from other sections
% [The comparison between the analytical data and empirical results demonstrated that the displacement simulated by the FEA closely mirrored the observed behavior of the tail joint. ]
% [From Table \ref{tab:data_table} , the variation between motors is small, validating that TALE is a repeatable test platform for biological hypothesis testing. \Zac {needs revision} ]
% \input{Sections/05_conclusion}

\bibliographystyle{ieeetr}
\bibliography{references}

\end{document}